\documentclass[10pt,twocolumn,letterpaper]{article}

\usepackage{lineno,hyperref}
\usepackage{colortbl}
\usepackage{amsmath,amssymb,amsfonts}
\usepackage{algorithmic}
\usepackage{graphicx}
\usepackage{float}

\usepackage[pagenumbers]{cvpr} 
\usepackage{colortbl}
\usepackage{multirow}

\definecolor{cvprblue}{rgb}{0.21,0.49,0.74}
\def\confName{CVPR}
\def\confYear{2025}

\newcommand{\method}{ByTheWay\xspace}
\newcommand{\specialfootnotetext}[1]{
  \begingroup
  \def\thefootnote{\fnsymbol{footnote}} 
  \footnotetext[0]{#1} 
  \endgroup
}

\title{ByTheWay: Boost Your Text-to-Video Generation Model to Higher Quality in a Training-free Way}

\author{
Jiazi Bu\textsuperscript{1,4*}
\quad
Pengyang Ling\textsuperscript{2,4*}
\quad
Pan Zhang\textsuperscript{4\dag}
\quad
Tong Wu\textsuperscript{3}
\quad
Xiaoyi Dong\textsuperscript{4}
\quad
Yuhang Zang\textsuperscript{4} \\ 
\quad
Yuhang Cao\textsuperscript{4} 
\quad
Dahua Lin\textsuperscript{3,4}
\quad
Jiaqi Wang\textsuperscript{4\dag}
\\
{\small \textsuperscript{1}SJTU, \textsuperscript{2}USTC, \textsuperscript{3}CUHK, \textsuperscript{4}Shanghai AI Laboratory}
}

\begin{document}

\twocolumn[
\maketitle
{
\renewcommand\twocolumn[1][]{#1}
\vspace{-3.5em}
\begin{center}
\centering
\captionsetup{type=figure}
\includegraphics[width=0.95\textwidth]{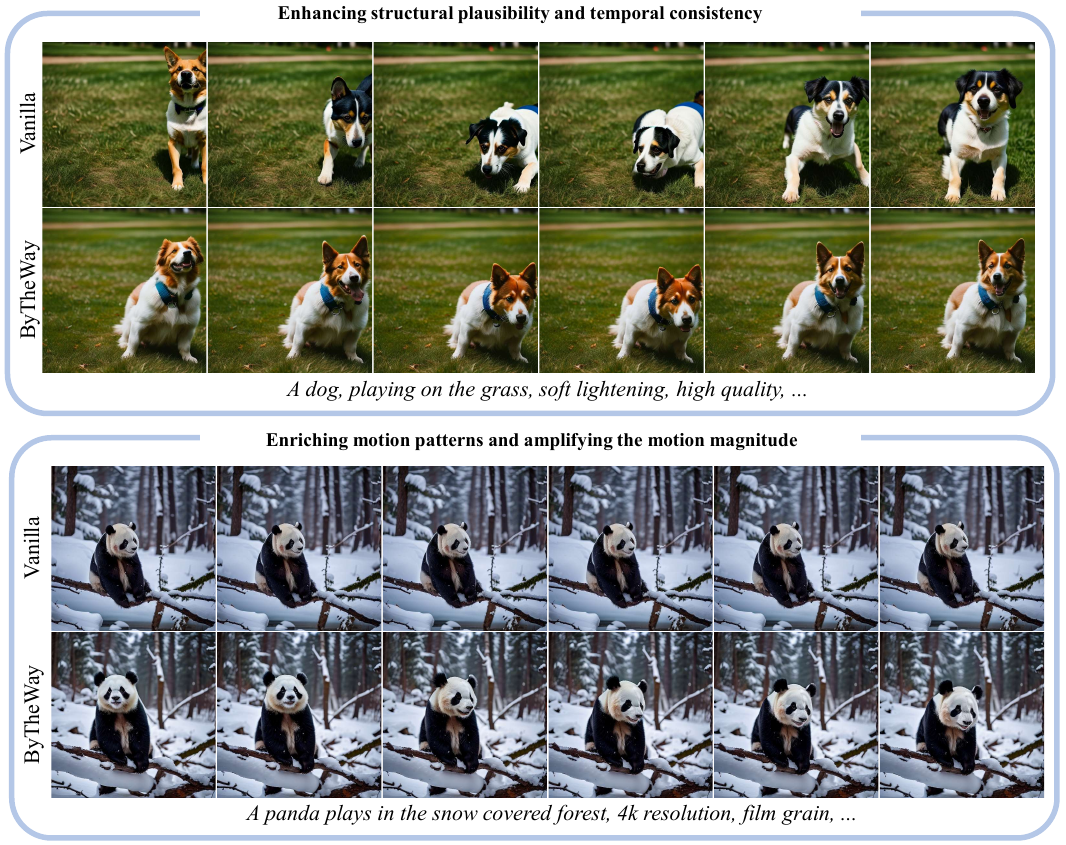}
\vspace{-0.5em}
\captionof{figure}{{\bf  Unlock the potential of pretrained text-to-video (T2V) generation models in a training-free approach.} (1) \method helps to enhance structural plausibility and temporal consistency in generated videos, significantly reducing artifacts and flickering. (2) \method contributes to enriching motion patterns and amplifying the motion magnitude in generated videos. Further, \method can be seamlessly integrated into various powerful T2V backbones (\eg, AnimateDiff\citep{guo2023animatediff} and VideoCrafter2\citep{chen2024videocrafter2}) in a plug-and-play manner, serving as a highly extensible module without introducing additional parameters or sampling cost.
}
\label{fig:teaser}
\end{center}
}]

\specialfootnotetext{* indicates equal contribution, \dag \ indicates corresponding author}
\begin{abstract}
The text-to-video (T2V) generation models, offering convenient visual creation, have recently garnered increasing attention. Despite their substantial potential, the generated videos may present artifacts, including structural implausibility, temporal inconsistency, and a lack of motion, often resulting in near-static video. In this work, we have identified a correlation between the disparity of temporal attention maps across different blocks and the occurrence of temporal inconsistencies. Additionally, we have observed that the energy contained within the temporal attention maps is directly related to the magnitude of motion amplitude in the generated videos. Based on these observations, we present \textbf{\method}, a training-free method to 
 improve the quality of text-to-video generation without introducing additional parameters, augmenting memory or sampling time. Specifically, \method\ is composed of two principal components: 1) Temporal Self-Guidance improves the structural plausibility and temporal consistency of generated videos by reducing the disparity between the temporal attention maps across various decoder blocks. 2) Fourier-based Motion Enhancement enhances the magnitude and richness of motion by amplifying the energy of the map.  Extensive experiments demonstrate that \method \ significantly improves the quality of text-to-video generation with negligible additional cost.
\end{abstract}    

\section{Introduction}
\label{sec:intro}

In recent years, the field has observed substantial progress in the evolution of diffusion-based models specifically dedicated to video generation tasks, notably in text-to-video synthesis~\citep{khachatryan2023text2videozero,blattmann2023alignyourlatents,guo2023animatediff,chen2024videocrafter2}. Despite these advancements, the practical applicability of generated videos remains limited due to inadequate quality. This suboptimal performance is characterized by two predominant issues: firstly, a portion of the generated videos exhibit structurally implausible and temporally inconsistent artifacts, and secondly, another subset of the generated videos demonstrates markedly restricted motion, bordering on the static nature of a still image. Prior methodologies have primarily concentrated on enhancing video generation quality through advances in training mechanisms, such as improving the quality of training data~\citep{blattmann2023stablevideodiffuion}, scaling training data~\citep{wang2024recipe_scaleup}, refining model architecture~\citep{hong2022cogvideo} and training strategies~\citep{chen2024videocrafter2}. However, these approaches often entail substantial costs. This work endeavors to improve video generation quality in the inference phase, specifically in the realm of text-to-video generation, without necessitating training, introducing additional parameters, augmenting memory or sampling time.

In current video generation models, an encoder-decoder architecture~\citep{ronneberger2015unet} is typically utilized, wherein the decoder is comprised of multiple blocks. Each block integrates several temporal attention modules~\citep{guo2023animatediff}, facilitating the modeling of motion within the generated videos. We have two observations about the temporal attention module. The first is a correlation between artifact presence and the inter-block divergence of temporal attention maps. Specifically, video generation processes exhibiting structurally implausible and temporally inconsistent artifacts demonstrate greater disparity between the temporal attention maps of different decoder blocks. Conversely, processes devoid of such evident artifacts exhibit reduced disparity among these maps, as illustrated in Fig. \ref{fig:statistic}(a). The second is a correlation between the amplitude of motion in generated videos and the energy of the corresponding temporal attention maps, defined in the method section. Specifically, videos that exhibit a higher degree of motion amplitude and a richer variety of motion patterns are observed to possess greater energy within their temporal attention maps, as illustrated in Fig. \ref{fig:statistic}(c). 

Based on the observations, we present \method, a training-free approach with negligible additional cost to improve the generation quality of T2V diffusion models. 
\method\ is composed of two principal components: Temporal Self-Guidance and Fourier-based Motion Enhancement, both meticulously engineered to refine the temporal attention module within T2V models. Temporal Self-Guidance leverages the temporal attention map from the preceding block to inform and regulate that of the current block. This approach effectively mitigates the disparity between the temporal attention maps across various decoder blocks, thereby normalizing their disparity. As a result, videos that initially exhibit structural implausibility and temporal inconsistency, significantly reduce such artifacts through the application of Temporal Self-Guidance, as shown in the first and second rows in Fig.~\ref{fig:teaser}.
Furthermore, Fourier-based Motion Enhancement modulates the high-frequency components of the temporal attention map, thereby amplifying the energy of the map, as detailed in the methodology section. This enhancement circumvents the generation of videos that closely resemble static image. With the Fourier-based Motion Enhancement, videos that were previously characterized by minimal motion exhibit an increased amplitude and a more diverse range of motion patterns, as illustrated in the third and last rows in Fig.~\ref{fig:teaser}.

We evaluate the performance of \method \ on various popular T2V backbones, including those with additional motion modules trained from frozen T2I models and those trained end-to-end directly for T2V. Our experiments show promising results, demonstrating the effectiveness and strong adaptability of \method. Moreover, experiments reveal that \method \ also exhibits potential in the image-to-video (I2V) domain, further expanding the applicability of \method \ across various video generation tasks.

Our contributions are summarized: (1) We conduct a deeper analysis of the temporal attention module widely adopted in T2V backbones, and observe two correlations between the generated videos and corresponding temporal attention maps. (2) We propose \method,  which significantly improves the quality of T2V generation without necessitating training, introducing additional parameters, augmenting memory or sampling time. (3) \method\ can be seamlessly integrated with various mainstream open-source T2V backbones like AnimateDiff and VideoCrafter2, showing strong applicability and extensibility.

\section{Related Work}
\label{sec:related}

\subsection{Text-to-Video Diffusion Models}
Given a textual prompt, text-to-video (T2V) diffusion models \citep{singer2022make_a_video,hong2022cogvideo,wang2023lavie,chen2023videocrafter1,wang2023modelscope,wang2024videocomposer,khachatryan2023text2videozero} aim to synthesize image sequences that maintain both temporal consistency and textual alignment. Unlike text-to-image~\citep{ding2021cogview,zeqiang2023minidalle,saharia2022photorealistic,podell2023sdxl} that emphasizes perfecting individual images, T2V poses a heightened challenge of maintaining both visual aesthetics for each frame and the realistic motion between frames. To this end, most approaches incorporate extra motion modeling modules into existing image diffusion architecture, leveraging the underlying image priors. For instance, AnimateDiff \citep{guo2023animatediff} introduced trainable temporal attention layers to frozen text-to-image models to effectively capture the frame-to-frame correlations.  Some works~\citep{blattmann2023alignyourlatents,chen2024videocrafter2} combined temporal convolution modules and temporal attention layers for modeling short/long range dependencies. To alleviate motion synthesis difficulty, Ge et al.~\citep{ge2023preserve_noiseprior} suggested employing temporally related noise to enhance temporally consistent. Nevertheless, due to the scarcity of high-quality video data and the intricacies of motion synthesis, the current available T2V models still struggle to harmonize motion strength with motion consistency. This work identifies that the consistency across temporal attention blocks indicates the continuity of synthesized video sequences while the energy within the temporal attention maps dominates the magnitude of motion, and thus proposes a training-free strategy to unlock the potential of exiting T2V models by encouraging uniform motion modeling and enhanced frequency energy.

\subsection{Diffusion Feature Control}
Controlling diffusion features to manipulate specific attributes has been demonstrated to be an effective strategy in the realm of image and video synthesis~\citep{chefer2023attendandexcite,kim2023denset2i,xiao2023fastcomposer,liu2023videop2p,qi2023fatezero}. Prompt2Prompt \citep{hertz2022prompt2prompt} revealed that the cross attention maps domain the image layout. DSG \citep{yang2023dsg} proposed that spatial means of diffusion features represent the appearance, which offers simple approach for image property manipulation, such as size, shape, and location.
FreeControl \citep{mo2023freecontrol} suggested to perform image structure guidance by aligning the PCA features with given reference image in spatial self-attention block, providing a versatile counterpart of ControlNet \citep{zhang2023controlnet}.
DIFT \citep{tang2023emergent} observed that the semantic correspondence can be directly extracted by spatially measuring the difference between diffusion feature.  FreeU \citep{si2023freeu} suggested re-weighting the contribution of skip features and backbone features by using spectral modulation and structure-related scaling, promoting the emphasis on backbone semantics. In the field of video generation, MotionClone \citep{ling2024motionclone} demonstrated the sparse control of temporal attention maps facilitates a training-free motion transfer, enabling reference-based video generation. FreeInit\citep{wu2025freeinit} proposed to alleviate the initialization gap in video generation by iteratively refining the low-frequency components of initial latent, but suffers from increased inference cost and attenuated motion.
UniCtrl\citep{chen2024unictrl} suggested to improve content alignment across frames by sharing the keys/values of first frame in self-attention layers, which produces reduced motion magnitude and thus requires extra branch for motion preservation. I4VGEN\citep{guo2024i4vgen} decomposed text-to-video generation in a sequential manner, which demands the collaboration of I2V models and is characterized by higher inference cost.
In this work, we propose Temporal Self-Guidance to facilitates uniform motion modeling across blocks by narrowing the disparities between temporal attention maps. This is work together with Fourier-based Motion Enhancement, which boosts motion magnitude by amplifying frequency energy, thus elevating the quality of the generated videos.
\section{Preliminary}
\label{sec:preliminary}

\subsection{Latent Diffusion Model} 

In the context of T2V generation, latent diffusion model~\citep{rombach2022high_sd} is widely as backbone as its significant advancement in image synthesizing. Typically, based on a pre-trained autoencoder $\mathcal{E}(\cdot)$ and $\mathcal{D}(\cdot)$, video sequences are projected into the latent space, in which a denoising network $\epsilon_\theta$ is encouraged to learn the mapping from noised video latent $z_t$ to pure video latent $z_0$. Mathematically, the noised video latent $z_t$ obeys the following distribution:
\begin{equation}\label{eq:noise_distribution}
z_t = \sqrt{\bar{\alpha}_t}z_0 + \sqrt{1-\bar{\alpha}_t}\epsilon,
\end{equation}
where $\bar{\alpha}_t$ is a pre-defined parameter representing noise schedule~\citep{ho2020ddpm}, $ \epsilon \sim \mathcal{N}(0, 1)$ is the added noise, and $t \sim \mathcal{U}(1, T)$ denotes time step. To restore $z_0$ from $z_t$, denoising network $\epsilon_\theta$ is forced to estimate the noise component in $z_t$, which can be expressed as:
\begin{equation}\label{eq:denoise_loss_raw}
\mathcal{L(\theta)} = \mathbb{E}_{z_0, \epsilon, t} \left[\| \epsilon_t - \epsilon_{\theta}(z_{t}, c, t) \|_{2}^{2}\right],
\end{equation}
where $c$ represents the textual prompt. During sampling, $z_t$ is initialized with Gaussian noise and undergoes iterative denoising conditioned on $c$ for prompt-aligned generation.

\begin{figure*}
\centering
\includegraphics[width=1.0\linewidth]{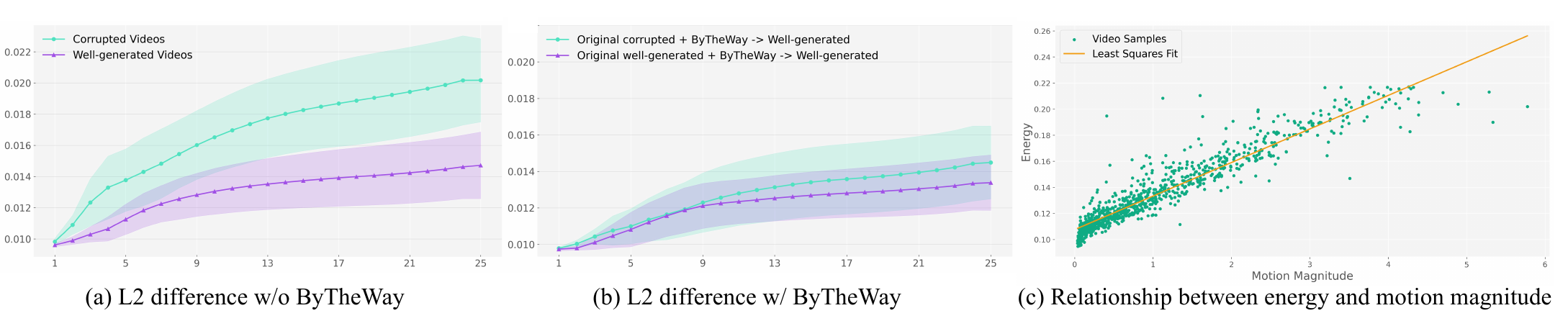}
\vspace{-2.5em}
\caption{\textbf{Statistical patterns derived from T2V generation process.} (a) Generated videos exhibiting structurally implausible and temporally inconsistent artifacts demonstrate greater disparity between the temporal attention maps of different decoder blocks. (b) After applying \method, the modeling disparity in original corrupted videos are reduced to the level of well-generated videos. (c) Videos with larger motion magnitude typically exhibit higher energy, in which the motion magnitude is measured by the estimated optical flow.}
\vspace{-0.8em}
\label{fig:statistic}
\end{figure*}

\begin{figure}
\centering
\includegraphics[width=\linewidth]{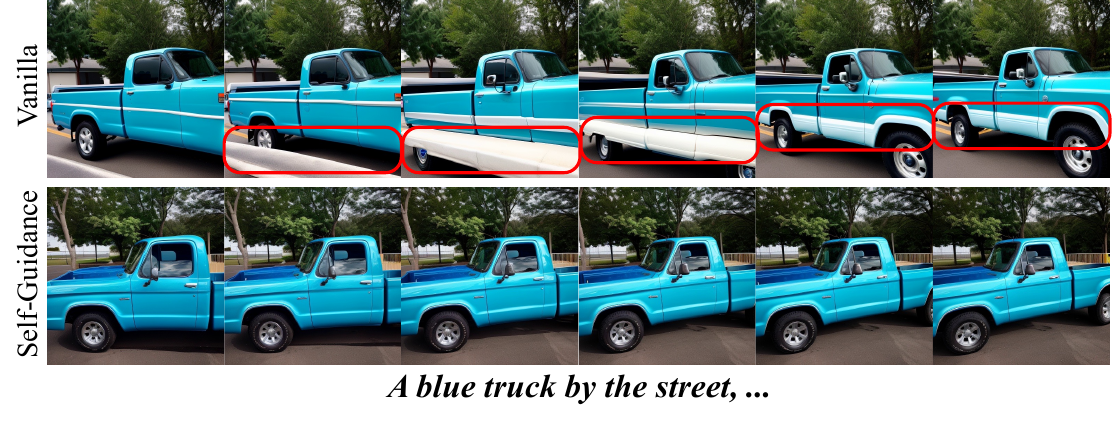}
\vspace{-2.5em}
\caption{{\bf Temporal Self-Guidance}. Temporal Self-Guidance contributes to the restoration of
collapsed structures and consistency of motion in the generated video.
}
\label{fig:self-guidance}
\end{figure}

\subsection{Temporal Attention Mechanism}
The biggest difference between video generation and image generation lies in the synthesis of motion, i.e., the modeling of correlation between video sequences. This is typically achieved by temporal attention mechanism, which establishes feature interactions across frames via self-attention operations in temporal dimension. For 5D video diffusion feature $f \in \mathbb{R}^{B\times C\times F \times H \times W}$, where $B$ and $F$ represent batch axis and frame time axis, $H$ and $W$ denotes spatial resolution,  temporal attention performs self-attention in its 3D reshaped variant $f' \in \mathbb{R}^{(B\times H \times W) \times C\times F }$, in which the generated attention map $\mathcal{A}\in\mathbb{R}^{(B\times H\times W) \times F \times F}$ reflects the temporal correlation between frames.

\section{Method}
\label{sec:method}

\subsection{Temporal Self-Guidance}

Temporal attention modules are extensively integrated at various hierarchical stages within the upsampling blocks of T2V architectures \citep{blattmann2023alignyourlatents, guo2023animatediff, chen2023videocrafter1, chen2024videocrafter2}. These modules, derived from different tiers of the diffusion U-Net, are employed to capture inter-frame dependencies at multiple resolutions. 
We conjecture that, due to the limited capability in modeling motion, the nearby temporal attention maps struggle to capture large motion increments, which can lead to implausible structures and temporal inconsistencies. To substantiate this hypothesis, we analyzed 100 structurally and motion-degraded videos alongside 100 well-generated videos.
The motion disparities are defined as the incremental differences in motion modeling across various blocks within the model, which are quantified by calculating the L2 difference between the temporal attention maps of the \texttt{up\_blocks.1} and the subsequent blocks. As illustrated in Fig.~\ref{fig:statistic}(a), it is observed that significant disparities between temporal attention maps across different blocks are associated with the occurrence of implausible structures and temporal inconsistencies.

To mitigate the excessive divergence between temporal attention maps across various upsampling blocks, we introduce a straightforward yet potent temporal self-guidance mechanism. This mechanism involves the infusion of the temporal attention map of \texttt{up\_blocks.1} into subsequent blocks, modulated by a guidance ratio $\alpha$. The adjustment can be mathematically modeled as:
\begin{equation}\label{eq:self-guidance}
    \mathcal{A}_m = \mathcal{A}_m + \alpha (\mathcal{A}_1^{m} - \mathcal{A}_m),
\end{equation}
where $\mathcal{A}_m$ denotes the temporal attention map of $m$-th upsampling block ($m = 2, 3$), and $\mathcal{A}_1^{m}$ refers to the temporal attention map of \texttt{up\_blocks.1}, which is upsampled to match the spatial dimensions of $\mathcal{A}_m$. As depicted in Fig.~\ref{fig:statistic} (b) and Fig.~\ref{fig:self-guidance}, the implementation of temporal self-guidance effectively alleviates the excessive modeling disparity between different hierarchical levels of temporal attention modules, thereby diminishing the structurally implausible and temporally inconsistent artifacts in the resultant video generation. 

Beyond addressing the structural implausibility and temporal inconsistency issues resolved by Temporal Self-Guidance, we have observed that some generated videos, including those corrected by Temporal Self-Guidance, still suffer from a paucity of motion, often appearing nearly static. To tackle this, we introduce a novel strategy aimed at amplifying the motion amplitude and diversity within the generated videos by capitalizing on the energy inherent in the temporal attention maps.

\subsection{Fourier-based Motion Enhancement}

\subsubsection{Energy Representation of Motion Magnitude} \label{sec:energy}

\begin{figure}
\centering
\includegraphics[width=\linewidth]{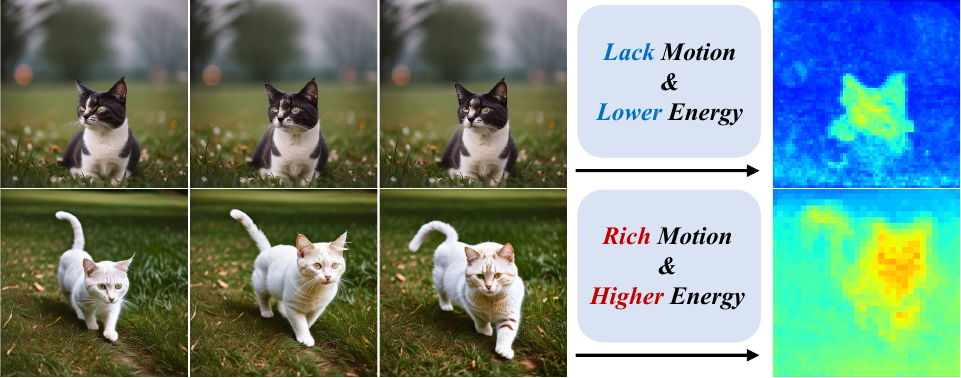}
\vspace{-1.8em}
\caption{\textbf{Energy representation of video motion magnitude}. Samples with richer motion typically
exhibit a higher energy.}
\label{fig:intro}
\end{figure}

The temporal attention map encapsulates a rich set of motion-related information that is pivotal for the generation of dynamic video content. We find that the energy encapsulated within the temporal attention map is indicative of the motion amplitude present in the generated video. To elaborate, consider a temporal attention map $\mathcal{A} \in \mathbb{R}^{(B \times H \times W) \times F \times F}$, where $B$ represents the batch size, $H \times W$ denotes the spatial resolution, and $F$ is the number of frames. The energy $E$ of this map can be quantified by the following equation:
\vspace{-0.5em}
\begin{equation}\label{eq:energy}
    E = \frac{1}{F} \sum_{i=0}^{F-1} \sum_{j=0}^{F-1}||\mathcal{A}_{..., i,j}||^2, 
\vspace{-0.5em}
\end{equation}
as illustrated in Fig.~\ref{fig:intro}. To substantiate the correlation between the energy of the temporal attention map and the motion magnitude in the generated video, we employ the RAFT \citep{teed2020raft} to extract the optical flow, using the average magnitude of this flow as a metric for motion strength. Our findings reveal a positive correlation: videos with greater motion magnitudes are associated with higher spatially averaged energies within their temporal attention maps, as depicted in Fig.~\ref{fig:statistic} (c). This insight motivates us to manipulate the motion magnitude in the generated videos by modulating the energy intensity of the temporal attention maps. By doing so, we aim to enhance the dynamism and variability of the motion in the videos.

\subsubsection{Motion Enhancement by Frequency Re-weighting} \label{sec:freq}

\begin{figure}
\centering
\includegraphics[width=\linewidth]{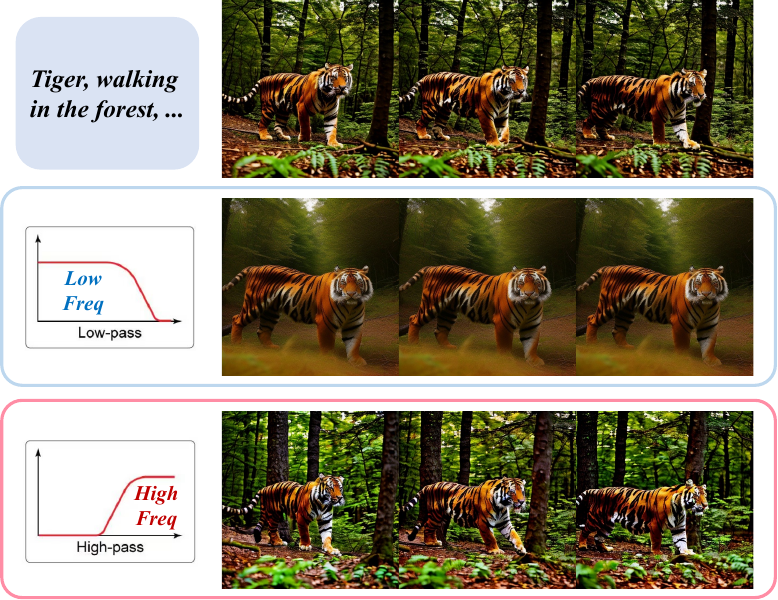}
\vspace{-2em}
\caption{
        {\bf Frequency decomposition.}
        By directly removing either the high-frequency or low-frequency components from the temporal attention map, it can be observed that motion in generated videos is primarily present in the high-frequency components.
}
\vspace{-1.2em}
\label{fig:freq}
\end{figure}

To enhance the motion amplitude in generated videos by amplifying the energy of the temporal attention map, we must overcome the challenge posed by the softmax normalization inherent in attention maps, which precludes straightforward numerical scaling. To address this, we employ a sequence-to-sequence discrete frequency decomposition technique, specifically the Fast Fourier Transform (FFT), to the temporal attention map. For a given temporal attention map $\mathcal{A} \in \mathbb{R}^{(B \times H \times W) \times F \times F}$, we decompose it into its high-frequency and low-frequency components as follows:
\begin{equation}
\begin{aligned}
\mathbf{A} &= \mathcal{F}(\mathcal{A}), \\
\mathbf{A}_H &= \mathbf{A}_{..., i_H}, ~i_H \in [\frac{F}{2} - \tau, \frac{F}{2} + \tau],\\
\mathbf{A}_L &= \mathbf{A}_{..., i_L}, ~i_L \in [0, \frac{F}{2} - \tau) \cup (\frac{F}{2} + \tau, F-1],
\end{aligned}
\end{equation}
where $\mathcal{F}$ denotes the 1D FFT operation along the softmax axis, $\mathbf{A} \in \mathbb{C}^{(B \times H \times W) \times F \times F}$ is the complex-valued matrix resulting from applying the FFT to $\mathcal{A}$, and $\tau$ is a hyperparameter that determines the frequency range for the high-pass and low-pass filters. As demonstrated in Fig.~\ref{fig:freq}, experiments involving the selective removal of high-frequency or low-frequency components from the temporal attention map during the denoising process have yielded insightful observations. Videos that retain only the low-frequency components tend to exhibit a nearly static structure, closely mirroring the characteristics of their unmodified counterparts. In contrast, videos that include solely high-frequency components display abundant motion but are marred by inconsistency and persistent flickering. These findings suggest that the essence of motion in generated videos is predominantly encapsulated within the high-frequency components of their temporal attention maps. 

Motivated by these insights, we introduce a scaling factor $\beta$ to modulate the high-frequency components $\mathbf{A}_H$. The process of scaling and reconstructing the temporal attention map is formalized by the following equation:
\begin{equation}\label{eq:idft}
    \mathcal{A}^{'} = \widetilde{\mathcal{F}}(\beta \mathbf{A}_H + \mathbf{A}_L),
\end{equation}
where $\widetilde{\mathcal{F}}$ represents the inverse Fast Fourier Transform (iFFT) operation, and $\mathcal{A}^{'}$ signifies the temporal attention map with the scaled high-frequency components. Based on aforementioned equations, we have the following theorems (detailed proof is provided in the supplementary material).

\textbf{Theorem 1.} \textit{For any $\beta \geq 0$, $\mathcal{A}^{'}$  possesses the softmax property. Specifically, $\sum_k \mathcal{A}^{'} = \sum_k \mathcal{A} = \mathbf{I}$, where $k$ denotes the softmax dimension associated with $\mathcal{A}$, and $\mathbf{I}$ is an all-ones matrix.}

Therefore, $\mathcal{A}^{'}$ can replace $\mathcal{A}$ as the new temporal attention map in the decoder.

\textbf{Theorem 2.} \textit{If  $\beta > 1$, then the energy of $\mathcal{A}^{'}$, denoted as $E_x^{'}$, is greater than the energy of $\mathcal{A}$, denoted as $E_x$. Conversely, if $0 < \beta < 1$, then $E_x^{'}$ is less than $E_x$.} 

Subsequently, the motion magnitude of generated videos can be enhanced with improved energy under $\beta > 1$.

\begin{figure}[htp]
\centering
\includegraphics[width=\linewidth]{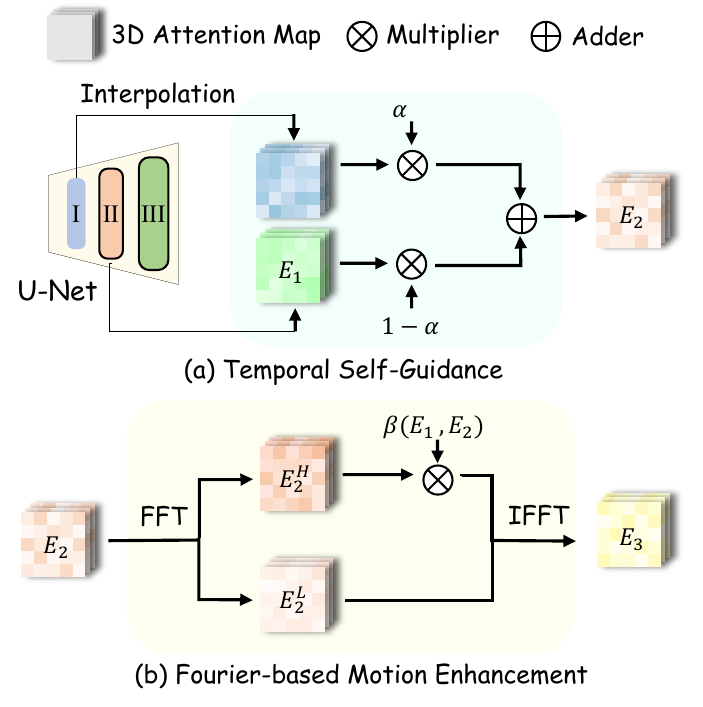}
\vspace{-2em}
\caption{
        {\bf \method Operations.}
        (a) Temporal Self-Guidance. The temporal attention map from \texttt{up\_blocks.1} is injected into the corresponding modules of \texttt{up\_blocks.2/3} with a guidance ratio $\alpha$, in order to enhance the structural plausibility and temporal consistency. (b) Fourier-based Motion Enhancement. A scaling factor $\beta$ is applied to the high-frequency components of the temporal attention map, thereby amplifying the motion magnitude within generated videos.
}
\vspace{-1em}
\label{fig:pipeline}
\end{figure}

\subsection{\method Operations}

Leveraging the techniques proposed above, we introduce \method, a training-free method to enhance the T2V quality without increasing inference expense. As illustrated in Fig.~\ref{fig:pipeline}, \method initially applies Temporal Self-Guidance to improve the structural coherence and temporal consistency of the video. Subsequently, Fourier-based Motion Enhancement is employed to amplify motion dynamics. 
To ensure that the motion magnitude of generated videos processed by \method exceeds that of the original, unenhanced videos, the energy of the temporal attention map after Fourier-based Motion Enhancement, denoted as $E_3$, must be greater than the energy of the original temporal attention map, denoted as $E_1$. To achieve this, the scaling factor $\beta$ is defined as a function of the energies before and after Temporal Self-Guidance, $E_1$ and $E_2$, respectively:
\begin{equation}\label{eq:beta}
\beta(E_1, E_2)=max\{\beta_0, \sqrt{\frac{E_1 - E_2^{L}}{E_2^{H}}}\},
\end{equation}
where $\beta_0$ is user-given value of $\beta$ to control the motion magnitude. $E_2^{H}$ and $E_2^{L}$ denoting the energies of the high-frequency and low-frequency of the attention map after applying Temporal Self-Guidance, respectively. See the supplementary material for a detailed explanation for Eq. \ref{eq:beta}.
\section{Experiments and Results}
\label{sec:experiments}

\begin{figure*}[htp]
\centering
\includegraphics[width=1.0\linewidth]{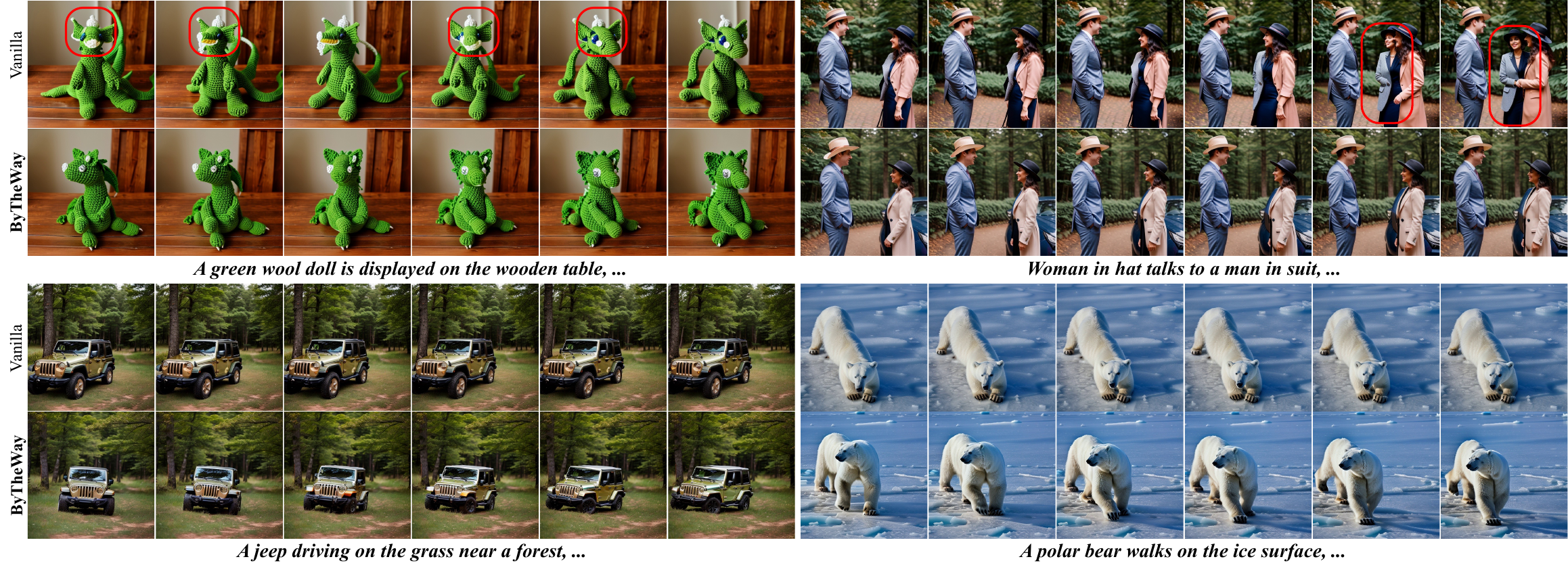}
\vspace{-2.5em}
\caption{\textbf{Samples generated by AnimateDiff\citep{guo2023animatediff} with or without \method}.}
\vspace{-1em}
\label{fig:comparison-animatediff}
\end{figure*}

\begin{figure*}[htp]
\centering
\includegraphics[width=1.0\linewidth]{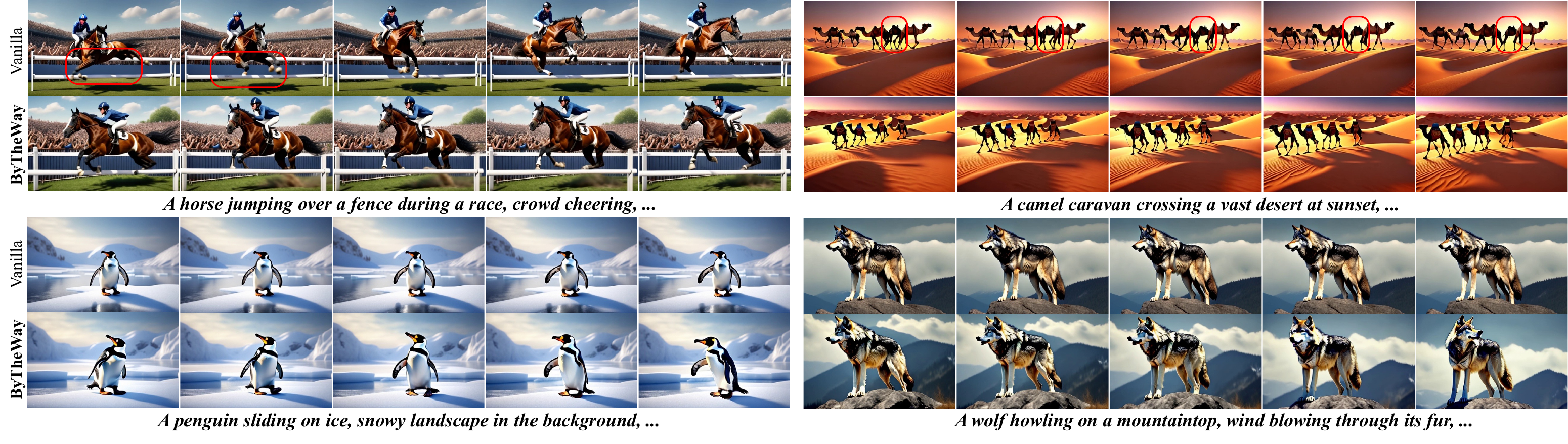}
\vspace{-2.5em}
\caption{\textbf{Samples generated by VideoCrafter2\citep{chen2024videocrafter2} with or without \method}.}
\vspace{-1.5em}
\label{fig:comparison-videocrafter2}
\end{figure*}

\subsection{Experiments Setup}

\textbf{Base models.} We mainly conduct our experiments on two mainstream T2V backbones with superior visual quality: AnimateDiff ($512\times 512$)~\citep{guo2023animatediff} with Realistic Vision V5.1 LoRA and VideoCrafter2 ($320\times 512$)~\citep{chen2024videocrafter2}. Results generated by vanilla backbones are used as a baseline. \method \ operations are only applied during the first 20\% steps of the denoising process. DDIM sampler~\citep{song2020ddim}
with classifier-free guidance~\citep{ho2022classifierguidance} is adopted in the inference phase.

\noindent \textbf{Parameter setup.} The \method \ parameters are set to $\alpha = 0.6$, $\beta = 1.5$, $\tau = 7$ in default for AnimateDiff, $\alpha = 0.1$, $\beta = 10$, $\tau = 7$ in default for VideoCrafter2. Note that the default values of \method parameters are relatively robust within a specific T2V backbone but may not be universally optimal for different backbones, since different base models exhibit variations in their motion preferences.

\noindent \textbf{Evaluation metrics.} We report three metrics for quantitative evaluation. First, we conduct a user study with 30 participants to assess \textit{Video Quality}, considering both structure coherence and motion magnitude. Secondly, we employ GPT-4o~\citep{achiam2023gpt4} for a comprehensive Multimodal-Large-Language-Model (MLLM) Assessment on hundreds of generated videos. The implementation details are available in the supplementary material. Moreover, we evaluate 200 videos generated by Vanilla T2V backbones and \method-enhanced backbones using VBench~\citep{huang2023vbench}.

\subsection{Qualitative Comparison} 

As presented in Fig.~\ref{fig:comparison-animatediff} and Fig.~\ref{fig:comparison-videocrafter2}, with the integration of \method, various T2V backbones demonstrates a notable performance improvement compared to their vanilla synthesis results. For instance, giving AnimateDiff the prompt ``\textit{a green wool doll is displayed on the wooden table.}'', \method \ enhances the structural consistency of the synthesized video, preventing the collapse of the doll's head and tail. Moreover, in the ``\textit{A jeep driving on the grass near a forest.}'' case, \method \ amplifies the dynamic effects of the scene, making the jeep exhibit more pronounced motion. For VideoCrafter2, when provided with the prompt ``\textit{A horse jumping over a fence during a race, crowd cheering.}'', \method \ reconstructs the structure of the rider and horse, addressing the issue of structural anomalies in the horse’s legs while enhancing the overall motion to appear more synchronized and aesthetically pleasing. In cases like ``\textit{A penguin sliding on ice, snowy landscape in the background.}'', \method \ preserves the original structural integrity while introducing richer, more dynamic motion to the scene. 

\begin{figure}
\centering
\includegraphics[width=\linewidth]{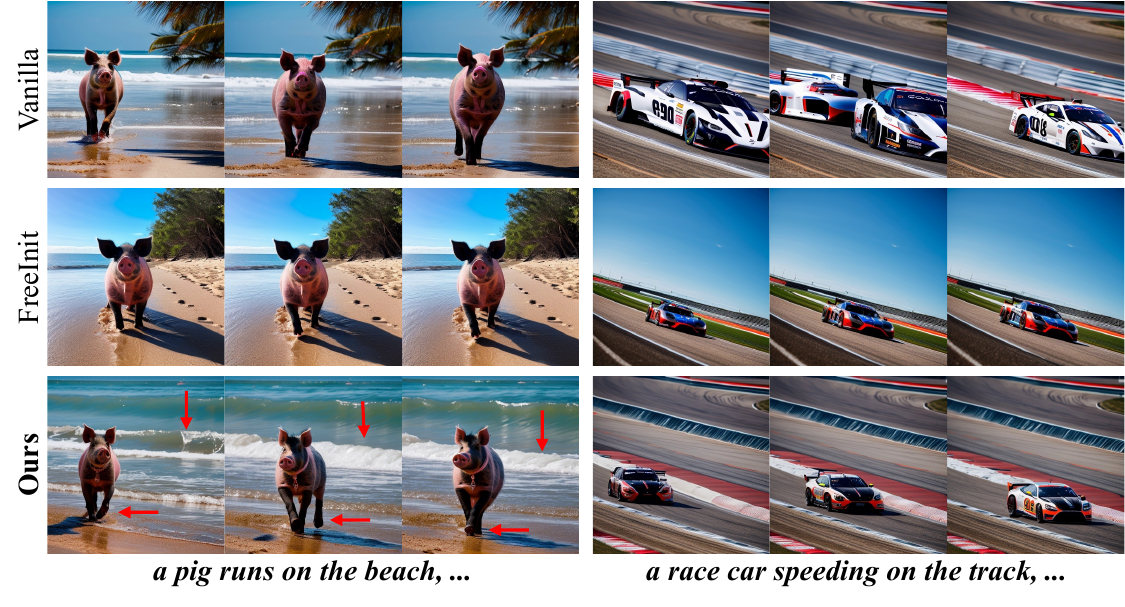}
\vspace{-2.5em}
\caption{
        {\bf Visual comparison with FreeInit~\citep{wu2025freeinit}}.
}
\vspace{-2.2em}
\label{fig:freeinit}
\end{figure}

FreeInit~\citep{wu2025freeinit} is a training-free method designed to improve T2V temporal consistency by iteratively refining the spatial-temporal low-frequency components of the initial latent throughout the denoising process. As shown in Fig.~\ref{fig:freeinit}, regardless of whether the vanilla-generated video is corrupted, FreeInit leads to a significant loss of motion while refining its structure, resulting in a nearly static video. In contrast, \method simultaneously enhances both structural coherence and motion magnitude, achieving a comprehensive improvement in video quality. Additionally, FreeInit requires iterative cycles to refine the initial noise, adding significant time overhead (5$\times$ sampling time), while \method incurs almost no extra inference cost.

In summary, \method \ effectively improves the structural consistency of synthesized videos while amplifying their motion dynamics, resulting in a significant enhancement in the overall synthesis quality of the T2V backbones.

\subsection{Quantitative Evaluation}

\textbf{User Study.} As shown in Table~\ref{tab:vote} (a),  \method receives the majority of votes, demonstrating its remarkable improvement in generating visually appealing results. 

\begin{table}[t]
\resizebox{\linewidth}{!}{
\begin{tabular}{lccc}
\toprule[1.5pt]
\multirow{2}{*}{Evaluation Metric} & \multicolumn{1}{c}{(a) User Study} & \multicolumn{2}{c}{(b) MLLM Assessment} \\ \cmidrule(r){2-2} \cmidrule(r){3-4}
   & Video Quality~ & Structure Rationality~  & Motion Consistency~  \\ \midrule[0.5pt]
AnimateDiff~\citep{guo2023animatediff}    & 25.42\%  & 41.94\% &  34.62\% \\
\rowcolor{gray!10}\textbf{+ \method}           & \textbf{74.58\%}  & \textbf{58.06\%} & \textbf{65.38\%}  \\
\addlinespace
VideoCrafter2~\citep{chen2024videocrafter2} & 30.54\%  & 18.48\% & 39.60\%   \\
\rowcolor{gray!10}\textbf{+ \method}   & \textbf{69.46\%}    & \textbf{81.52\%} & \textbf{60.40\%}  \\ 
 \bottomrule[1.5pt]
\end{tabular}
}
\centering
\vspace{-1em}
\caption{\textbf{Voting results of user study and MLLM assessment.} }
\vspace{-1em}
\label{tab:vote}
\end{table}

\begin{table*}[]
\resizebox{\linewidth}{!}
{
\begin{tabular}{lccccccc}
\toprule[1.5pt]
\multirow{2}{*}{Evaluation Metric} &  \multicolumn{6}{c}{VBench~\citep{huang2023vbench} Metrics} \\  \cmidrule(r){2-7} 
   & Subject Consistency~$\uparrow$  & Background Consistency~$\uparrow$ & Motion Smoothness~$\uparrow$ & Dynamic Degree~$\uparrow$  & Aesthetic Quality~$\uparrow$ & Imaging Quality~$\uparrow$ \\ \midrule[0.5pt]
AnimateDiff~\citep{guo2023animatediff}    & 0.9318 & 0.9507 & 0.9474    & 0.4073 & 0.6376 & 0.7215   \\
AnimateDiff + FreeInit~\citep{wu2025freeinit}            & 0.9712 & 0.9695 & 0.9713    & 0.2941 & 0.6571 & 0.7459 \\
\rowcolor{gray!10}\textbf{AnimateDiff + \method}            & \textbf{0.9744} & \textbf{0.9725} & \textbf{0.9786}    & \textbf{0.5245} & \textbf{0.6609} & \textbf{0.7553}  \\
\addlinespace
VideoCrafter2~\citep{chen2024videocrafter2}  & 0.9732 & 0.9691 & 0.9749  & 0.4086 & 0.6477 & 0.6575   \\
\rowcolor{gray!10}\textbf{VideoCrafter2 + \method}            & \textbf{0.9852} & \textbf{0.9896} & \textbf{0.9863}  &  \textbf{0.5547} & \textbf{0.6598} & \textbf{0.6738}  \\ 
 \bottomrule[1.5pt]
\end{tabular}
}
\centering
\vspace{-1em}
\caption{\textbf{Quantitative results of \method on VBench~\citep{huang2023vbench}.}
\method facilitates the best performance of different T2V models.
}
\vspace{-1em}
\label{tab:vbench}
\end{table*}

\begin{figure*}
\centering
\includegraphics[width=\linewidth]{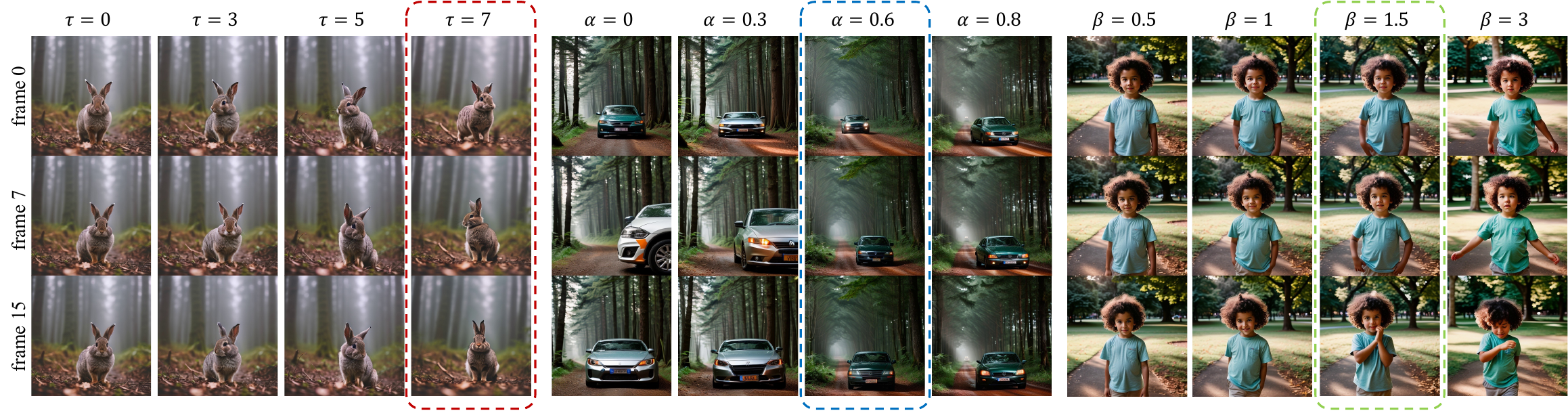}
\vspace{-2em}
\caption{
        {\bf Ablation on \method parameters.}
        Left: ``\textit{close up photo of a rabbit, forest, haze, ...}"; Middle:   ``\textit{car runs in the forest, ...}"; Right: ``\textit{Kid with curly hair plays in the park, ...}". Dashed boxes indicates the optimal parameters we chose in the experiments.
}
\vspace{-1.5em}
\label{fig:ablation}
\end{figure*}

\noindent \textbf{MLLM Assessment.} In light of the impressive strides made by Multimodal-Large-Language-Models (MLLM) recently in image/video understanding, the state-of-the-art MLLM, i.e., GPT-4o~\citep{achiam2023gpt4}, is employed for video quality assessment, covering structural rationality and motion consistency. As shown in Table~\ref{tab:vote}(b), \method exhibits notable gains in both structure rationality and motion consistency, validating its role in substantial video quality enhancement.

\noindent \textbf{VBench Metrics.} To objectively evaluate the overall generation quality, VBench~\citep{huang2023vbench} is introduced to serve as a comprehensive benchmark. As presented in Table~\ref{tab:vbench}, \method shows substantial improvements in VBench metrics, indicates its efficacy in enhancing T2V generation quality. Moreover, it can be observed that \method outperforms FreeInit~\citep{wu2025freeinit}, which is a strong training-free solution, across all evaluated dimensions. 

\subsection{Image-to-Video}

\begin{figure}
\centering
\includegraphics[width=\linewidth]{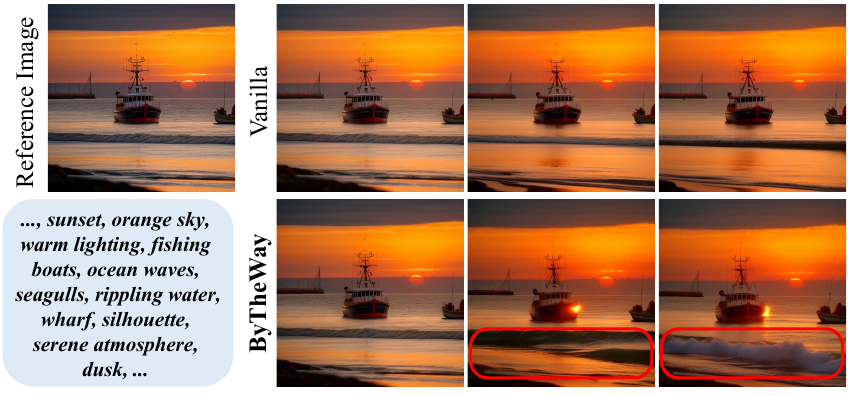}
\vspace{-2em}
\caption{
        {\bf Results of SparseCtrl~\citep{guo2023sparsectrl} with/without \method}.
}
\vspace{-2em}
\label{fig:i2v}
\end{figure}

Similar to text-to-video (T2V) tasks, image-to-video (I2V) is also a significant research area within video diffusion models. Here we employ SparseCtrl~\citep{guo2023sparsectrl}, a strong and flexible structure control method, as the I2V backbone to preliminarily validate the potential of \method \ in image-to-video tasks. As illustrated in Fig.~\ref{fig:i2v}, the infusion of \method \ into SparseCtrl serves to enhance the dynamic effects of the synthesized video while preserving the structural integrity of the reference image. Specifically, we observe that the video synthesized with \method \ exhibits more vivid wave motions, and the reflections of the setting sun display enhanced dynamic aesthetics. 

These experimental results demonstrate that \method \ effectively enhances the quality of both T2V and I2V generation tasks, positioning it as a versatile and powerful booster for video diffusion models.

\subsection{Ablation Study and Analysis}

\noindent \textbf{Effect of $\alpha$.} $\alpha$ represents the infusion ratio of lower-level temporal attention information in Temporal Self-Guidance. As shown in Fig.~\ref{fig:ablation}, an appropriate $\alpha$ strengthens the temporal consistency of the video, but an excessively large $\alpha$ may lead to the loss of motion information.

\noindent \textbf{Effect of $\beta$.} $\beta$ stands for the scaling factor of the high-frequency components in the temporal attention map within Fourier-based Motion Enhancement. An appropriate $\beta$ introduces richer and more intensified motion to the video, but an excessively large $\beta$ may cause the emergence of unexpected motion artifacts, as can be observed in Fig.~\ref{fig:ablation}. 

\noindent \textbf{Effect of $\tau$.} $\tau$ denotes the number of discrete frequency components involved in Fourier-based Motion Enhancement. As shown in Fig.~\ref{fig:ablation}, a larger $\tau$ allows for the manipulation of more frequency components that encode motion, thus promoting the motion enhancement effect.

\section{Conclusion}
\label{sec:conclusion}

In this work, we present \method, a training-free method to improve the quality of video generation without introducing additional parameters, augmenting memory or sampling time. \method consists of two key components: Temporal Self-Guidance and Fourier-based Motion Enhancement. The former improves structural plausibility and temporal consistency by reducing the disparity between temporal attention maps of different hierarchical levels. The latter enhances motion magnitude by scaling the high frequency of temporal attention maps. The proposed method can be easily integrated with available T2V backbones in a plug-and-play manner, offering a general and effective solution to enhance video generation quality during the inference phase.

\clearpage
{
    \small
    \bibliographystyle{main}
    \bibliography{main}
}

\clearpage
\setcounter{page}{1}
\maketitlesupplementary

In the supplementary material, we present additional qualitative results (Section~\ref{sec:add-qualitative}), more ablation experiments (Section~\ref{sec:add-ablation}), details of our user study and MLLM assessment (Section \ref{sec:add-quant}), the proof of Fourier-based Motion Enhancement (Section~\ref{sec:add-fourier}), as well as the limitation of our method (Section~\ref{sec:add-limitation}), as a supplement to the main paper. 

\section{Additional Qualitative Results}\label{sec:add-qualitative}

\begin{figure*}[htp]
\centering
\includegraphics[width=\linewidth]{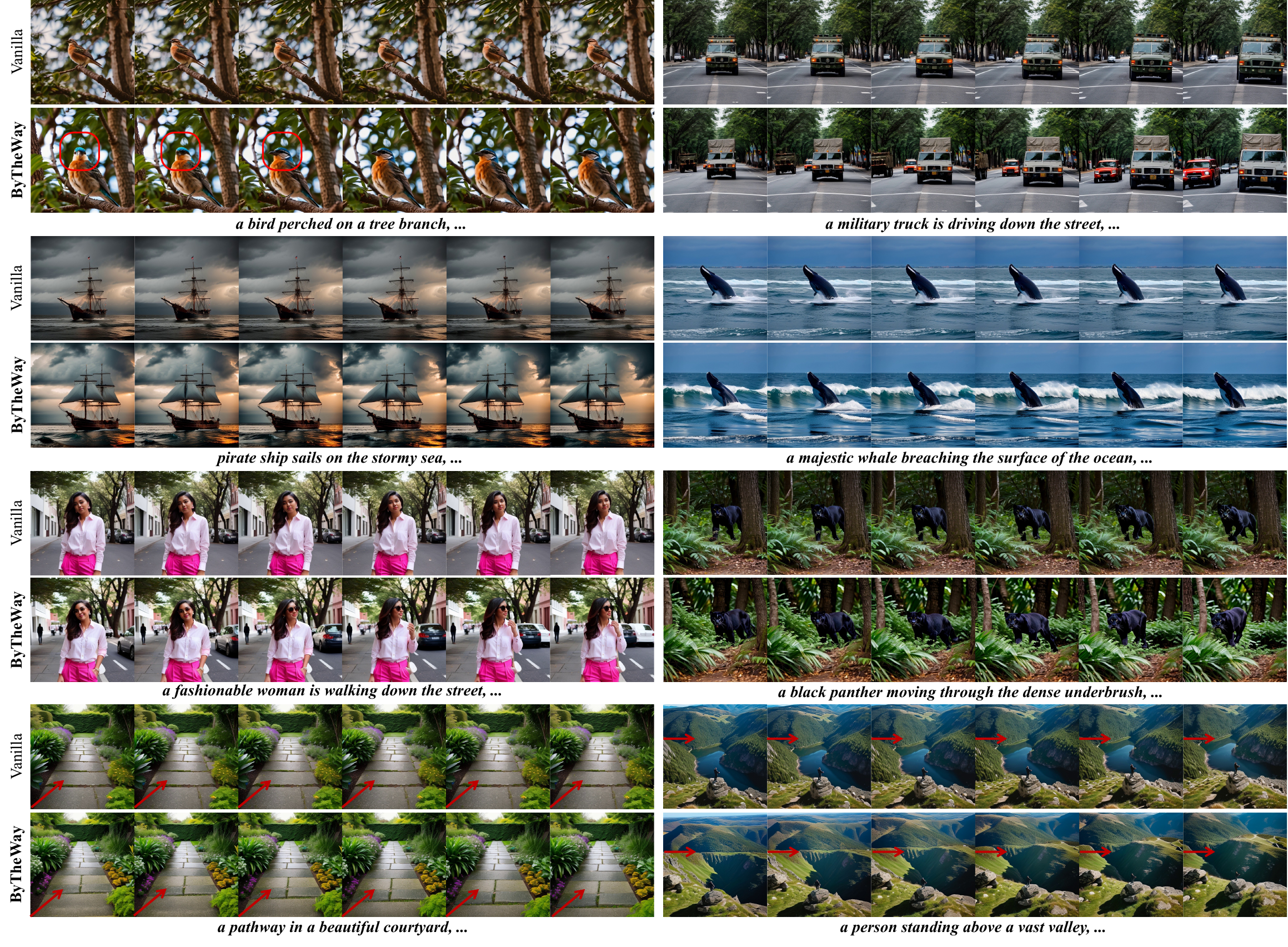}
\vspace{-2em}
\caption{
        {\bf More Results on AnimateDiff (Motion Enhancement).}
}
\vspace{-1em}
\label{fig:animatediff-motion}
\end{figure*}

\begin{figure*}[htp]
\centering
\includegraphics[width=\linewidth]{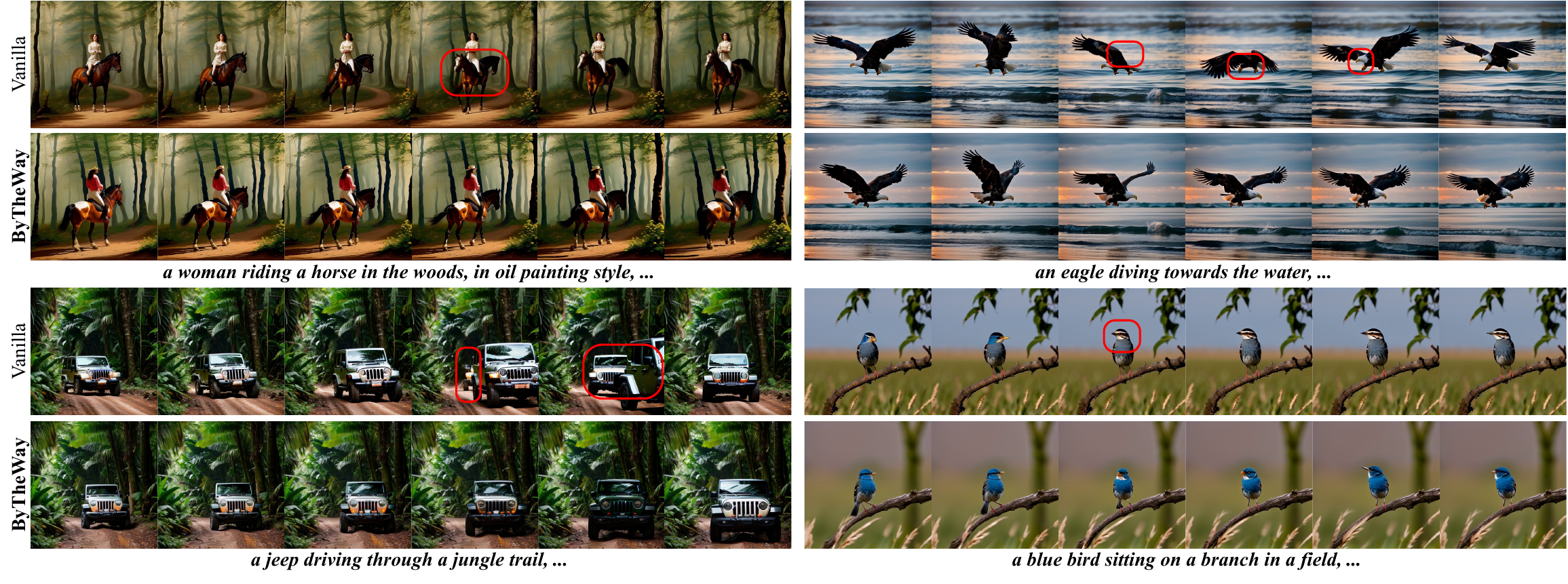}
\vspace{-2em}
\caption{
        {\bf More Results on AnimateDiff (Structure Enhancement).}
}
\vspace{-1em}
\label{fig:animatediff-consistency}
\end{figure*}

\textbf{More Results on AnimateDiff.} We present more results for video motion enhancement (Fig.~\ref{fig:animatediff-motion}) and structure enhancement (Fig.~\ref{fig:animatediff-consistency}) on AnimateDiff.

\begin{figure*}[htp]
\centering
\includegraphics[width=\linewidth]{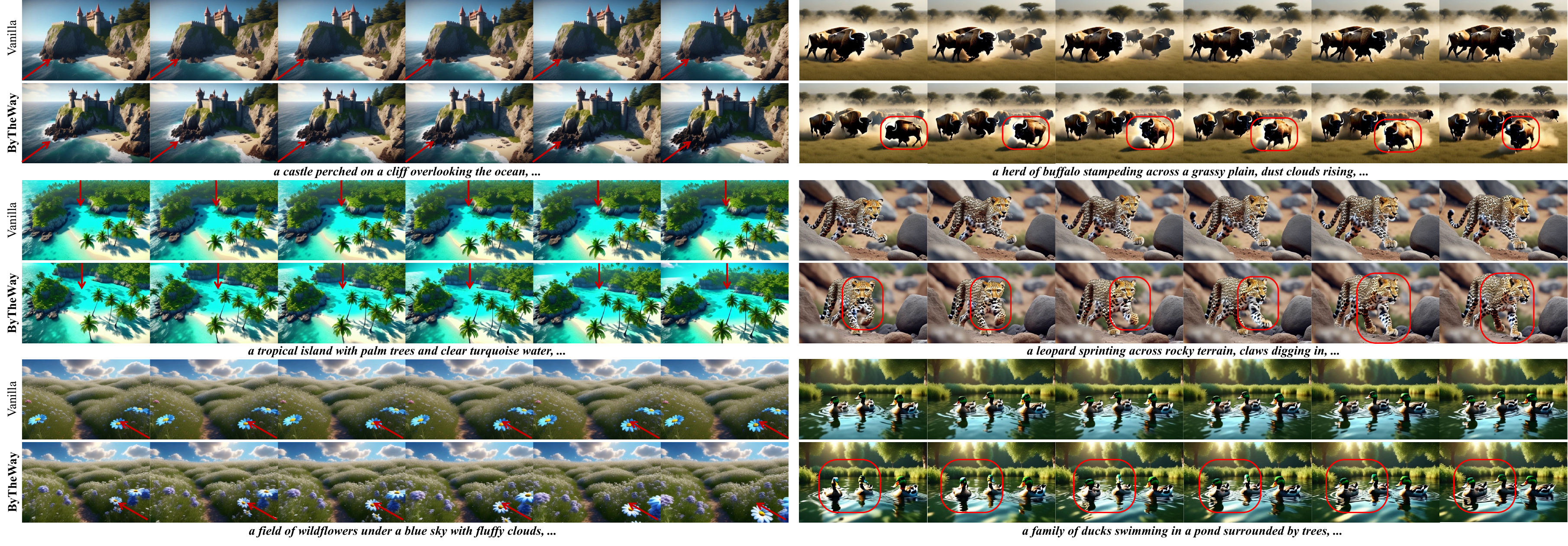}
\vspace{-2em}
\caption{
        {\bf More Results on VideoCrafter2 (Motion Enhancement).}
}
\vspace{-1em}
\label{fig:videocrafter2-motion}
\end{figure*}

\begin{figure*}[htp]
\centering
\includegraphics[width=\linewidth]{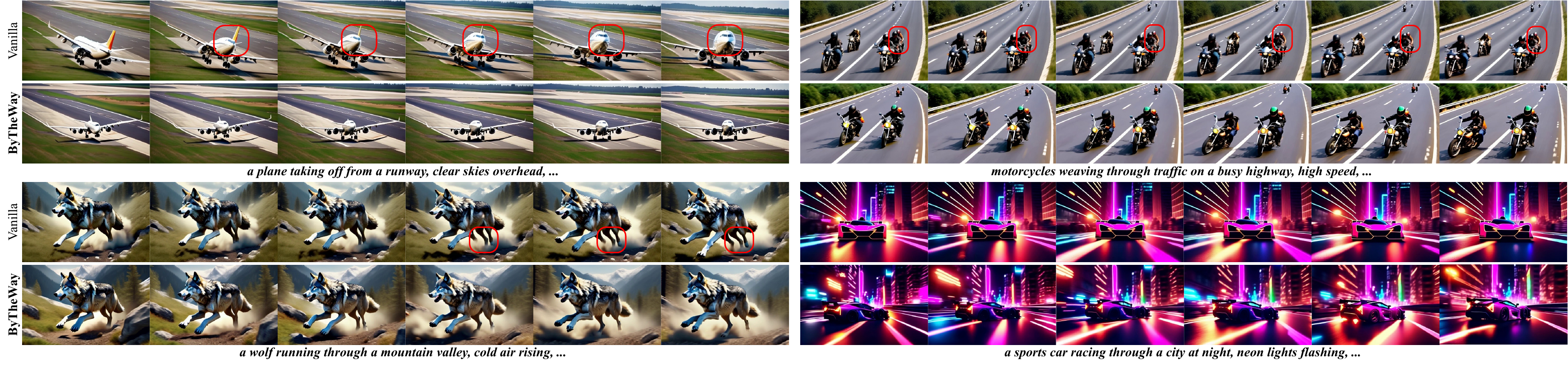}
\vspace{-2em}
\caption{
        {\bf More Results on VideoCrafter2 (Structure Enhancement).}
}
\vspace{-1em}
\label{fig:videocrafter2-consistency}
\end{figure*}

\noindent \textbf{More Results on VideoCrafter2.} We present more results for video motion enhancement (Fig.~\ref{fig:videocrafter2-motion}) and structure enhancement (Fig.~\ref{fig:videocrafter2-consistency}) on VideoCrafter2.

\begin{figure*}[htp]
\centering
\includegraphics[width=\linewidth]{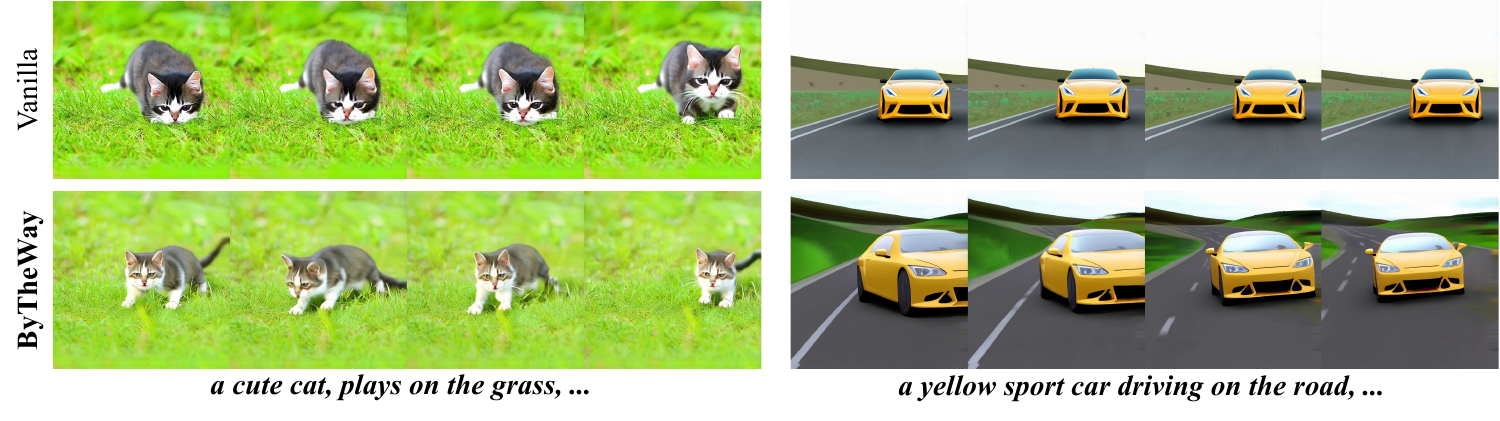}
\vspace{-2em}
\caption{
        {\bf Results on CogVideoX.}
}
\vspace{-1em}
\label{fig:dit}
\end{figure*}

\noindent \textbf{Application on DiT-based Architecture.} We extend \method to a DiT-based T2V backbone, CogVideoX. As depicted in Fig.~\ref{fig:dit}, \method showcases potential for motion enhancement in diffusion DiT.

\section{Additional Ablation Study}\label{sec:add-ablation}

\begin{figure}[htp]
\centering
\includegraphics[width=\linewidth]{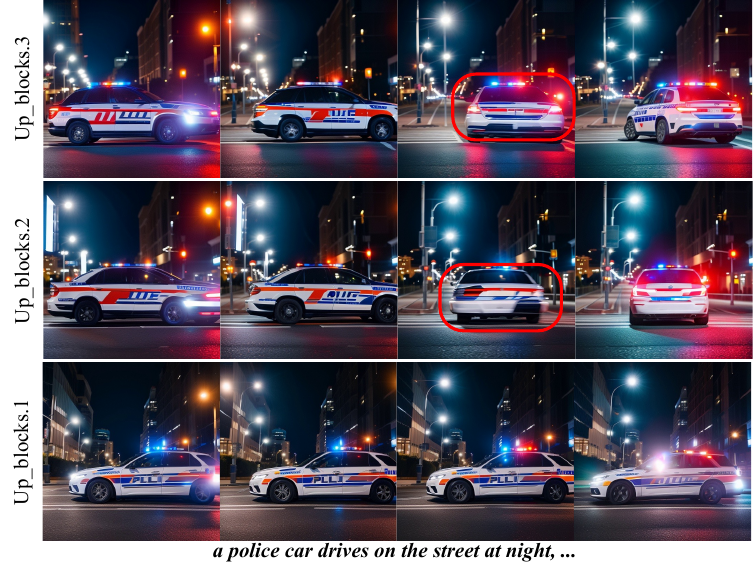}
\vspace{-2em}
\caption{
        {\bf Ablation on Guidance Anchor.}
}
\vspace{-1em}
\label{fig:ablation-anchor}
\end{figure}

\textbf{Choice of Guidance Anchor.} Fig.~\ref{fig:ablation-anchor} demonstrates that \texttt{up\_blocks.1} is the bottleneck in video motion modeling, injecting the temporal attention map of \texttt{up\_blocks.1} into subsequent decoder blocks helps align motion modeling across different levels of diffusion U-Net, thus enhance temporal consistency. In contrast, injecting information from later blocks fails to achieve this goal. Note that using \texttt{up\_blocks.3} as the anchor implies the absence of Temporal Self-Guidance (vanilla result).

\begin{figure*}[htp]
\centering
\includegraphics[width=\linewidth]{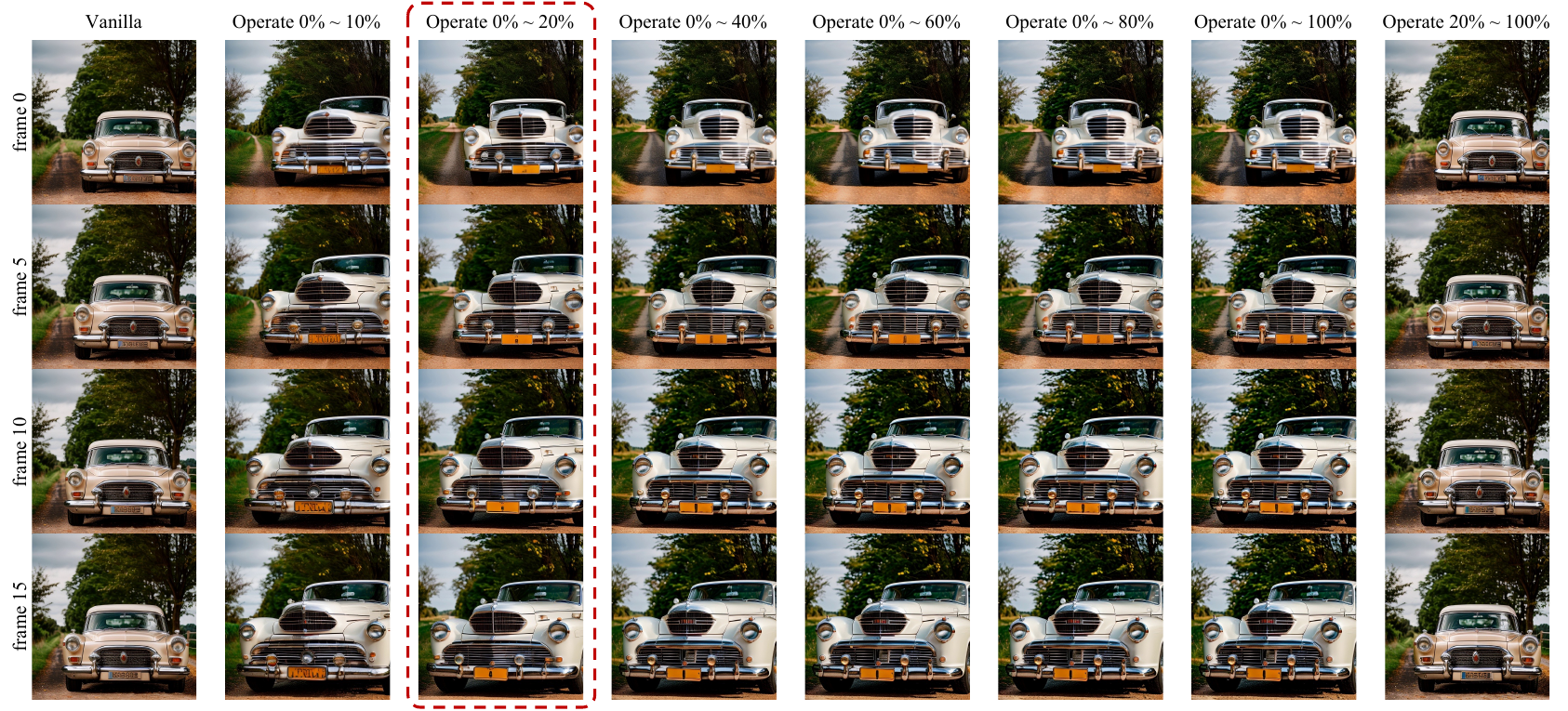}
\vspace{-2em}
\caption{
        {\bf Ablation on Operation Steps.} Prompt: \textit{``a vintage car drives on a country road, ..."}
}
\vspace{-1em}
\label{fig:ablation-step}
\end{figure*}

\noindent\textbf{Number of Operation Steps.} Fig.~\ref{fig:ablation-step} reveals that  the initial $20\%$ of sampling steps play a crucial role in shaping video motion, making the application of \method operations beyond this point have minimal effect on the generation quality. Moreover, when \method operations are applied only during $20\%$ to $80\%$ of the sampling steps, the generated video appears almost identical to the original video, which can be attributed that video motion is mainly determined by the early denoising stage. 

\begin{figure}[htp]
\centering
\includegraphics[width=\linewidth]{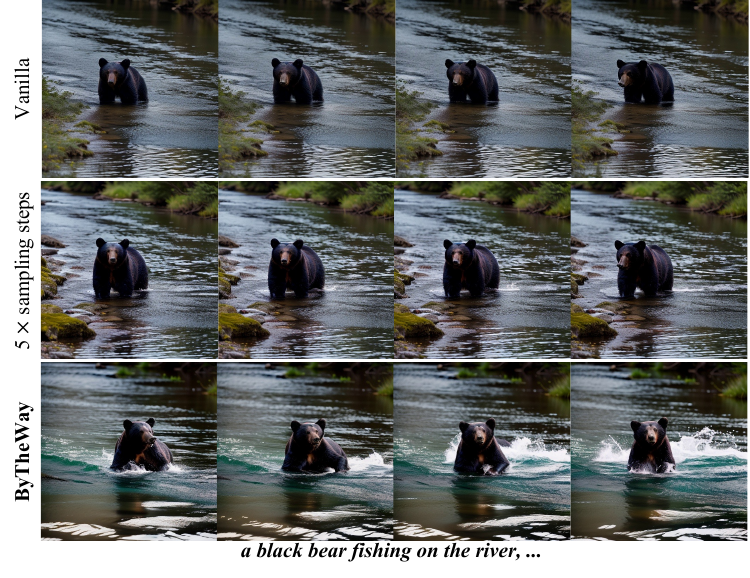}
\vspace{-2em}
\caption{
        {\bf Ablation on More Sampling Steps.}
}
\vspace{-1em}
\label{fig:ablation-more-step}
\end{figure}

\noindent\textbf{Does More Sampling Steps Help?} As shown in Fig.~\ref{fig:ablation-more-step}, the vanilla T2V backbone with $5\times$ sampling steps is inferior to incorporating the \method-enhanced backbone with only $1\times$ sampling steps, this demonstrates that \method is not equivalent to simply increasing the DDIM sampling steps.

\section{Details of User Study \& MLLM Assessment}\label{sec:add-quant}

\textbf{User Study Details.} In our user study, each participant receives 50 videos synthesized by Vanilla T2V backbones and 50 videos synthesized by \method-enhanced backbones. These videos are sampled from the same random seeds to ensure fair comparison. For each video pair from Vanilla and Vanilla + \method, participants are required to select the video they perceive as superior based on overall \textit{Video Quality}, considering both structure coherence and motion magnitude, and cast their vote accordingly. The videos were presented in a randomized order to reduce potential bias, and participants were allowed ample time to review each pair before making their selections.

\noindent\textbf{MLLM Assessment Prompt.} Here, we present the prompt used in the MLLM assessment.

\texttt{"""}

\textit{You are provided with two sets of video 
frames, each containing 4 representative 
frames, along with a shared textual prompt 
that was used to generate both videos. 
Your task is to perform a comparative 
evaluation of the two videos, focusing on their 
structure rationality / motion consistency.}

\textit{Here is the frame data of Video\_1.}

\textit{Here is the frame data of Video\_2.}

\textit{Based on your evaluation of motion consistency, choose the video 
set you find to be superior. 
If you determine that the first set of frames (Video\_1) is better, 
respond with "A". If the second set (Video\_2) is superior, respond 
with "B". Return only "A" or "B" based on your assessment.}

\texttt{"""}

\section{Fourier-based Motion Enhancement Proof}\label{sec:add-fourier}

In this section, we provide a detailed proof of how Fourier-based Motion Enhancement alters the energy of the temporal attention map in \method operations.

\subsection{Frequency Components Manipulation} \label{sec:manipulation}

Given a temporal attention map $\mathcal{A} \in \mathbb{R}^{(B\times H\times W) \times F \times F}$ with batch size $B$, spatial resolution $H \times W$ and frame number $F$, since we treat it as a batch of 1D attention sequences, we will next discuss the operations performed on a single softmax sequence $x[n]$ of length $F$. 

Mathematically, the operation of mapping the sequence $x[n]$ to the frequency domain is performed by the Discrete Fourier Transform (DFT):
\begin{equation}\label{eq:dft-definition}
    X[k] = \sum_{n=0}^{F-1} x[n] \cdot e^{-j \frac{2\pi}{N} kn},~k = 0, 1, \dots, F-1.
\end{equation}

Parseval's theorem states that the energy of a sequence is preserved under frequency domain transformation, meaning that the energy $E_x$ of sequence $x[n]$ is the same in both the time and frequency domains. This theorem can be expressed as follows:
\begin{equation}\label{eq:parseval}
    E_x = \sum_{n=0}^{F-1}x[n]^2 = \frac{1}{F} \sum_{k=0}^{F-1}X[k]^2.
\end{equation}

As mentioned in the main paper, Fourier-based Motion Enhancement uses a threshold index $\tau$ to separate the high-frequency and low-frequency components of the sequence, scaling the high-frequency components by a factor of $\beta$. This operation can be expressed as:
\begin{equation}\label{eq:fme}
X^{'}[k]=\left\{
\begin{array}{rcl}
\beta \cdot X[k]       &      & {k      \in    [\frac{F}{2}-\tau, \frac{F}{2} + \tau]},\\
X[k]     &      & {\textit{otherwise}},
\end{array} \right.
\end{equation}
after applying this manipulation, the energy $E_x^{'}$ of current attention sequence 
$x^{'}[n]$ is given by: 
\begin{equation}
    E_x^{'} = \frac{1}{F}[\sum_{k\notin [\frac{F}{2}-\tau, \frac{F}{2} + \tau]} X^{2}[k] + \beta^2\sum_{k\in [\frac{F}{2}-\tau, \frac{F}{2} + \tau]} X^{2}[k]],
\end{equation}
thus the energy change amount $\Delta E$ caused by Fourier-based Motion Enhancement can be computed as: 
\begin{align*}
    \Delta E &= E_x^{'} - E_x \\
             &= \frac{(\beta^2-1)}{F}\sum_{k\in [\frac{F}{2}-\tau, \frac{F}{2} + \tau]} X^{2}[k].
\end{align*}

Clearly, in the scenario where $\beta > 1$, Fourier-based Motion Enhancement will lead to an increase in the energy of the attention sequence ($\Delta E > 0$), while the opposite will result in a decrease in energy ($\Delta E < 0$), which elucidates the mechanism by which Fourier-based Motion Enhancement effectively enhances motion magnitude in synthesized videos.

Furthermore, it can be demonstrated that the attention sequence processed by Fourier-based Motion Enhancement remains a softmax sequence. This property is preserved because the direct current (DC) component $X[0]$ of the attention sequence, which determines the sum of the sequence, is not modified throughout the operation. By plugging $k=0$ into Eq. \ref{eq:dft-definition}, we can ascertain this property:
\begin{equation}
    X[0] = \sum_{n=0}^{F-1} x[n] = \sum_{n=0}^{F-1} x^{'}[n] = 1.
\end{equation}

\subsection{Adaptive $\beta$ in \method Operations} \label{sec:adaptive}

\begin{figure}[htp]
\centering
\includegraphics[width=\linewidth]{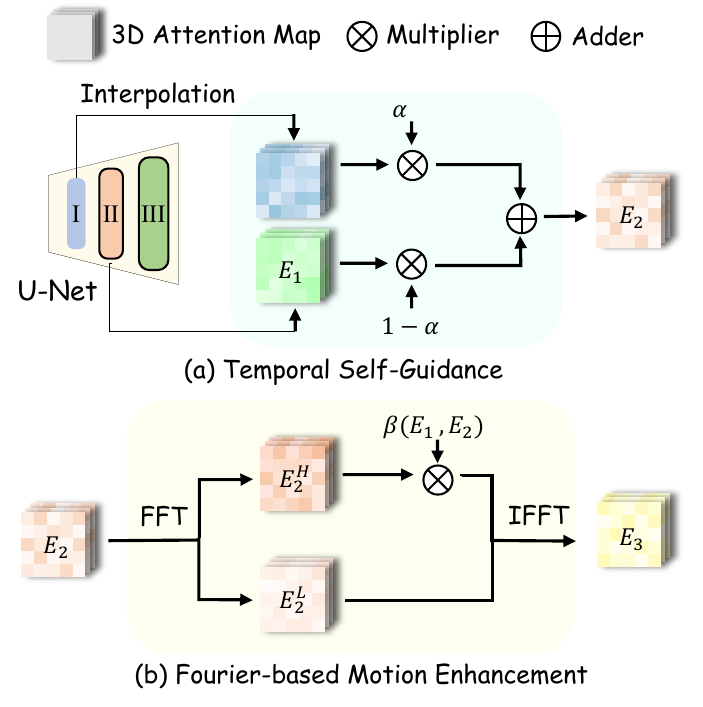}
\vspace{-2em}
\caption{
        {\bf \method Operations.}
}
\vspace{-1em}
\label{fig:add-pipeline}
\end{figure}

As depicted in the Fig.~\ref{fig:add-pipeline}, let $E_1$ denote the the energy of the temporal attention map before applying \method \ operations, $E_2$ the energy after Temporal Self-Guidance, and $E_3$ the energy after Fourier-based Motion Enhancement. Here, we demonstrate that using the adaptive $\beta$ as defined in Eq. \ref{eq:add-beta} ensures that $E_3\geq E_1$.
\begin{equation}\label{eq:add-beta}
\beta(E_1, E_2)=max\{\beta_0, \sqrt{\frac{E_1 - E_2^{L}}{E_2^{H}}}\},
\end{equation}

Based on the separation of high-frequency and low-frequency components in the sequence as described in Section \ref{sec:manipulation}, we can compute the energy of the high-frequency and low-frequency parts of the sequence $x[n]$, denoted as $E_x^H$ and $E_x^L$, respectively: 
\begin{equation}\label{eq:separation}
\begin{aligned}
E_x^H &= \frac{1}{F}\sum_{k\in [\frac{F}{2}-\tau, \frac{F}{2} + \tau]}X^2[k],\\
E_x^H &= \frac{1}{F}\sum_{k\notin [\frac{F}{2}-\tau, \frac{F}{2} + \tau]}X^2[k].
\end{aligned}
\end{equation}

According to Eq. \ref{eq:parseval} and Eq. \ref{eq:separation}, it is evident that the following relationship holds:
\begin{equation}\label{eq:high-low}
    E_x = E_x^H + E_x^L.
\end{equation}

Furthermore, we can concisely express the energy manipulation performed by Fourier-based Motion Enhancement described in Section \ref{sec:manipulation}, as follows:
\begin{equation}
    E^{'}_x = \beta^2 E_x^H + E_x^L,
\end{equation}
which indicates:
\begin{equation}
    E_3 = \beta^2 E_2^H + E_2^L.
\end{equation}

Therefore, to ensure $E_3\geq E_1$, it is necessary to ensure that $\beta$ adheres to the following condition: 
\begin{equation}\label{eq:beta-cond}
    \beta^2 E_2^H + E_2^L \geq E_1,
\end{equation}
the critical value of $\beta$, denoted as $\beta_c$, that satisfies this condition is:
\begin{equation}\label{eq:beta-critical}
    \beta_c = \sqrt{\frac{E_1 - E_2^L}{E_2^H}}.
\end{equation}

In \method operations, the user-specified $\beta$, denoted as $\beta_0$, will be compared with the critical value $\beta_c$, and the larger of the two will be selected as the actual $\beta$ value in Fourier-based Motion Enhancement:
\begin{equation}\label{eq:beta-max}
    \beta = max\{\beta_0, \beta_c\}.
\end{equation}

By adopting such a adaptive $\beta$ value, it can be theoretically guaranteed that the energy of the temporal attention map is increased during \method operations, thereby enhancing the motion magnitude in synthesized videos.

\section{Limitation}\label{sec:add-limitation}

Although \method demonstrates the capability to unlock the synthesis potential of various T2V backbones, the synthesized videos remain confined within the sampling distribution of the original T2V backbone. Therefore, one limitation of our method is that its performance upper bound is still constrained by the original T2V backbone.

\newpage


\end{document}